\documentclass[letterpaper]{article}
\usepackage{aaai24}
\usepackage{times}
\usepackage{helvet}
\usepackage{courier}
\usepackage[hyphens]{url}
\usepackage{graphicx}
\usepackage{natbib}
\usepackage{caption}
\usepackage{algorithm}
\usepackage{algorithmic}
\usepackage{bm}
\usepackage{amssymb}
\usepackage{amsmath}
\usepackage{array}
\usepackage{multirow}
\usepackage{booktabs}
\usepackage{pifont}
\usepackage[table]{xcolor}
\usepackage{color}
\definecolor{Gray}{gray}{0.93}
\usepackage{tikz}

\usepackage{newfloat}
\usepackage{listings}
\DeclareCaptionStyle{ruled}{labelfont=normalfont,labelsep=colon,strut=off}
\floatstyle{ruled}
\newfloat{listing}{tb}{lst}{}
\floatname{listing}{Listing}

\title{Compound Text-Guided Prompt Tuning via Image-Adaptive Cues}
\author{
    Hao Tan\textsuperscript{\rm 1,\rm 2}\equalcontrib,
    Jun Li\textsuperscript{\rm 1,\rm 2}\equalcontrib,
    Yizhuang Zhou\textsuperscript{\rm 3},
    Jun Wan\textsuperscript{\rm 1,\rm2}\thanks{Corresponding author},
    Zhen Lei\textsuperscript{\rm 1,\rm2,\rm4},
    Xiangyu Zhang\textsuperscript{\rm 3}
}
\affiliations{
    \textsuperscript{\rm 1}MAIS, Institute of Automation, Chinese Academy of Sciences, Beijing, China\\
    \textsuperscript{\rm 2}School of Artificial Intelligence, University of Chinese Academy of Sciences, Beijing, China\\
    \textsuperscript{\rm 3}MEGVII Technology\\
    \textsuperscript{\rm 4}CAIR, HKISI, Chinese Academy of Sciences, Hong Kong, China\\
    \{tanhao2023, lijun2021, jun.wan\}@ia.ac.cn, \{zhouyizhuang, zhangxiangyu\}@megvii.com, \\zlei@nlpr.ia.ac.cn
}

\begin{document}

\maketitle

\begin{abstract}
Vision-Language Models (VLMs) such as CLIP have demonstrated remarkable generalization capabilities to downstream tasks.
However, existing prompt tuning based frameworks need to parallelize learnable textual inputs for all categories, suffering from massive GPU memory consumption when there is a large number of categories in the target dataset. 
Moreover, previous works require to include category names within prompts, exhibiting subpar performance when dealing with ambiguous category names.
To address these shortcomings, we propose Compound \textbf{T}ext-\textbf{G}uided \textbf{P}rompt \textbf{T}uning (TGP-T) that significantly reduces resource demand while achieving superior performance.
We introduce text supervision to the optimization of prompts, which enables two benefits: 1) releasing the model reliance on the pre-defined category names during inference, thereby enabling more flexible prompt generation; 2) reducing the number of inputs to the text encoder, which decreases GPU memory consumption significantly.
Specifically, we found that compound text supervisions, i.e., category-wise and content-wise, is highly effective, since they provide inter-class separability and capture intra-class variations, respectively.
Moreover, we condition the prompt generation on visual features through a module called Bonder, which facilitates the alignment between prompts and visual features.
Extensive experiments on few-shot recognition and domain generalization demonstrate that TGP-T achieves superior performance with consistently lower training costs.
It reduces GPU memory usage by 93\% and attains a 2.5\% performance gain on 16-shot ImageNet.
The code is available at \url{https://github.com/EricTan7/TGP-T}.
\end{abstract}

\section{Introduction}
Large-scale vision-language pre-training~\cite{kim2021vilt,radford2021learning,jia2021scaling,bao2022vlmo} has emerged as a powerful paradigm for tackling a wide range of visual tasks~\cite{gu2021open,saharia2022photorealistic,alayrac2022flamingo}. 
The vision-language models (VLMs), e.g., CLIP~\cite{radford2021learning} and ALIGN~\cite{jia2021scaling}, have demonstrated remarkable generalization capabilities to various downstream tasks~\cite{yao2021cpt, guo2023texts, huang2023vop, smith2023coda}.
Among them, CLIP utilized the contrastive paradigm to align two modalities with 400 million image-text pairs.
One of its significant advancements is the ability to achieve open-vocabulary recognition by calculating the similarity between the query image feature and hand-craft text prompts (e.g., ``a photo of a $<$class$>$.'')
without additional training.

\begin{figure}[t]
	\centering
	\includegraphics[width=1.0\linewidth]{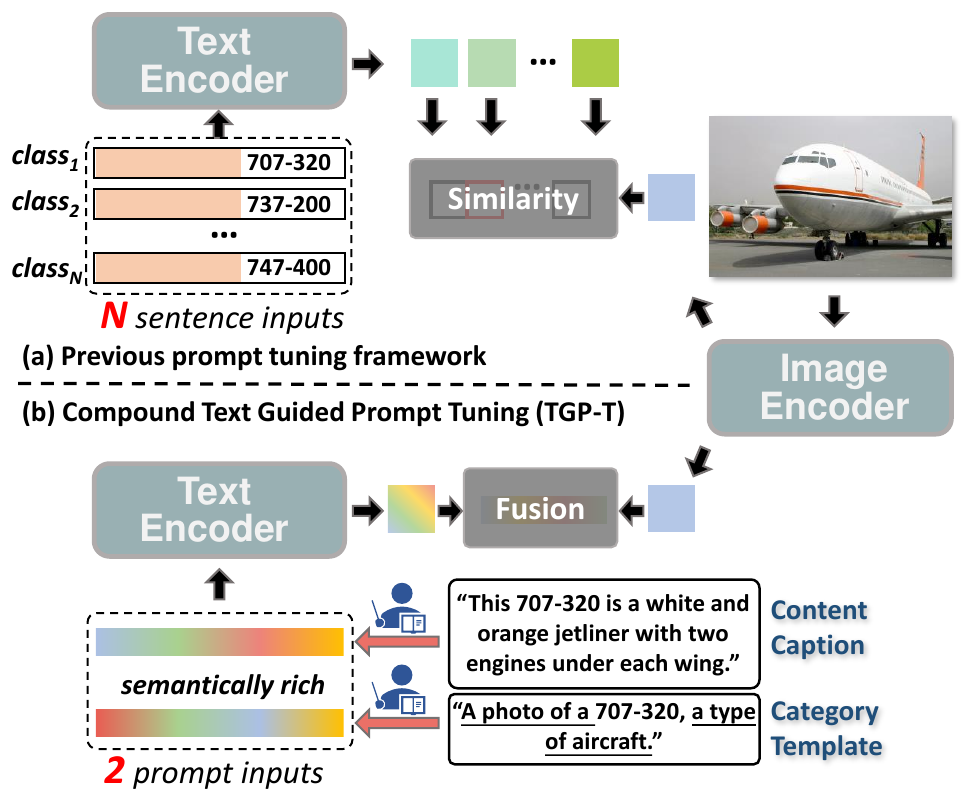}
	\caption{\textbf{Paradigm Comparison.} (a) Prior works parallelize $N$ learnable sentence inputs to text encoder and concatenate category names to each input. (b) TGP-T introduces text supervision to the optimization of prompts, which releases the reliance on category names and reduces the number of prompt inputs to two. This decreases GPU memory consumption significantly. 
 By employing two levels of text supervision, TGP-T performs strong adaptation ability.}
	\label{fig:intro}
\end{figure}

Fine-tuning a foundation model can be computationally expensive in the era of foundation models.
Consequently, there has been a shift in focus from adapting the model to specific tasks (i.e., fine-tuning) to fitting the downstream tasks with the foundation model (i.e., prompting)~\cite{liu2023pre}.
The milestone work CoOp~\cite{zhou2022learning} is the first time to apply prompt tuning to the VLMs, introducing a series of learnable prompts for textual input instead of taking hand-crafted templates. 
The follow-up CoCoOp~\cite{zhou2022conditional} uses visual features to construct adaptive prompts and further improve the generalization abilities.
Regarding data-scarce scenarios, e.g., few-shot recognition, prompt tuning has shown considerable improvements compared with manually curated text input.

\begin{figure}[t]
	\centering
	\includegraphics[width=1.0\linewidth]{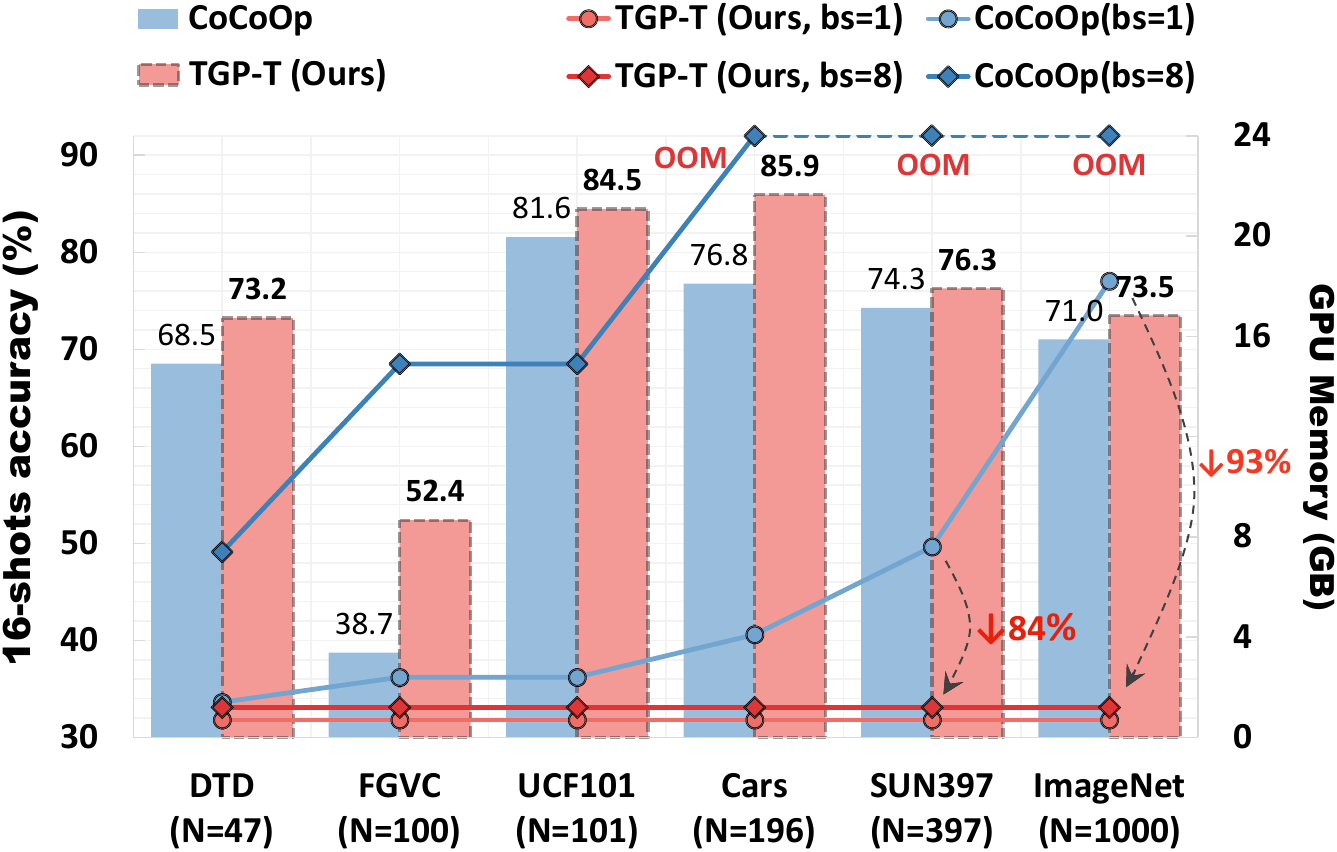}
	\caption{\textbf{Comparison on Performance (\%) and GPU Memory Consumption (GB).} $N$ is the number of categories and “bs" donates batch size. TGP-T reduces GPU memory usage across all datasets while achieving superior performance. The GPU memory consumption of TGP-T is independent of the number of categories. Note that when using batch size of 8, CoCoOp runs into out-of-memory (OOM) problems on StanfordCars, SUN397, and ImageNet with Nvidia RTX 3090.}
	\label{fig:mem}
\end{figure}

Despite notable progress, existing methods still suffer from certain shortcomings.
The series of methods~\cite{zhou2022learning, zhou2022conditional, zhu2022prompt,khattak2023maple, yao2023visual} based on CoOp simultaneously feed $N$ (the number of categories) learnable sentence inputs to text encoder.
This is equivalent to learning $N$ category centers before the text encoder, which requires preserving all intermediate activations of the text encoder for gradient backpropagation.
As a result, this approach leads to a rapid increase in GPU memory consumption as the $N$ grows, as shown by the blue curve in Fig.~\ref{fig:mem}.
Such a training process deviates from the original intention of efficient prompt tuning.
Based on the observation above, we propose to utilize a projector to \textit{relocate the learning of $N$ category centers after the text encoder.}
Accordingly, we only require two instead of $N$ prompts as the input to the text encoder.
As shown in Fig.~\ref{fig:mem}, these designs significantly reduce GPU memory consumption while achieving better performance.

Furthermore, prior works encounter another significant challenge.
The prompt tokens are prepended to each $<$class$>$ token.
Consequently, the classification weights, i.e., the textual features, are dependent on a pre-defined category name set, leading to subpar performance when dealing with ambiguous category names, e.g., ``707-320'' in the FGVCAircraft dataset~\cite{maji2013fine}. In this case, CoCoOp only achieves 38.7\% accuracy on 16-shot FGVCAircraft.
In contrast, we \textit{avoid including any category name within prompts}.
As shown in Fig.~\ref{fig:mem}, our method obtains a 13.7\% improvement on FGVCAircraft, exhibiting superior potential when addressing ambiguous category names.

To sum up, we propose a novel approach called Compound \textbf{T}ext-\textbf{G}uided \textbf{P}rompt \textbf{T}uning (TGP-T) that releases the huge resource demand while achieving state-of-the-art performance. 
As illustrated in Fig.~\ref{fig:intro}: 
1) we avoid including category names within prompts during inference, which allows for a more flexible prompt generation. 
2) Instead of parallelizing $\bm{N}$ \textbf{prompt inputs} for each image, we only require \textbf{two prompt inputs} to text encoder.
This reduces GPU memory consumption by 93\% while achieving a 2.5\% increase in accuracy on the 16-shot ImageNet, as shown in Fig.~\ref{fig:mem}.
Moreover, the GPU memory consumption of our framework is almost unaffected by the number of categories, which makes it accessible to tune VLMs on datasets of any scale. 

However, a crucial problem is to ensure the prompts carry sufficient information that is closely related to the current sample, without directly including category names.
Specifically, we suppose that \textit{the optimization of prompts should not be unconstrained}.
Therefore, we introduce two ``teachers'' to guide the process as shown in Fig.~\ref{fig:intro}.
1) \textit{Category-wise Text Supervision}, which offers a high-level understanding of the target category.
2) \textit{Content-wise Text Supervision}, which captures the intra-class variations.
The two ``teachers'' provide general category cues and specific content details, helping the model adapt to varying degrees of sample diversity and complexity.
Moreover, to incorporate fine-grained visual cues, we condition the prompt generation on the visual features~\cite{zhou2022conditional, rao2022denseclip} through a well-designed structure called Bonder.
During inference, we simply input the image into the network, and the Bonder generates suitable prompts via image-adaptive cues.
Then the prompts are fed into the text encoder to get enriched textual features.
Finally, the visual features and textual features are concatenated together, and then a projector is applied to obtain the predicted probability of the input image.
Our main contributions are summarized as follows:
\begin{itemize}
    \item We propose a novel framework called Compound \textbf{T}ext-\textbf{G}uided \textbf{P}rompt \textbf{T}uning (TGP-T), which significantly reduces training costs while achieving state-of-the-art performance on 11 datasets for few-shot classification.
    \item We offer an alternative perspective for prompt tuning, i.e., using text supervision to guide the optimization of prompts. This enables two benefits: 1) releasing the model reliance on category names during inference, and thereby enabling more flexible prompt generation; 
    2) reducing the number of inputs to the text encoder, which decreases GPU memory consumption significantly.
    \item Through empirical study, we found that compound~text supervisions, i.e., \textit{category-wise} and \textit{content-wise}, are highly effective. Since they provide inter-class separability \mbox{and capture intra-class variations, respectively.} 
    \item We propose to use a lightweight structure called Bonder to bridge the visual and language modalities.
    By interacting prompt queries with image features, the Bonder facilitates the generated prompts to be closely aligned with the current image features, which allows \mbox{a better harness of VLM.}
\end{itemize}

\section{Related Works}

\textbf{Vision-Language Models.}
There has been a growing interest in vision-language models since CLIP~\cite{radford2021learning} was proposed.
With a contrastive-based pretraining approach, CLIP has achieved impressive progress in visual representation learning by utilizing large-scale image-text pairs.
Many fields have benefited from CLIP, such as object detection~\cite{gu2021open}, image generation~\cite{saharia2022photorealistic}, and visual question answering~\cite{alayrac2022flamingo}.
With the rapid development of large language models (LLM)~\cite{brown2020language,touvron2023llama} in recent times, it has become possible to directly apply separately pre-trained LLM and vision foundation models to build VLM, which is capable of understanding both images and text by adding a small number of connection parameters for training.
For instance, BLIP-2~\cite{li2023blip} achieves this by training the additional Q-former and linear mapping.
Similarly, MiniGPT-4~\cite{zhu2023minigpt} and LLaVA~\cite{liu2023visual} achieve impressive multimodal chat abilities with only additional training of linear projection \mbox{for aligning the two modalities.}

Benefiting from the advancements of these VLMs, we further explored how to utilize CLIP to improve visual recognition tasks. 
Inspired by~\cite{li2023blip}, we propose to use a module to bridge the vision and language modalities, which facilitates the learnable prompts to be closely aligned with the visual features.
In addition, by leveraging these VLMs to generate text descriptions for images, we inject rich cross-modal knowledge into the vision tasks.
\\
\\
\noindent\textbf{Prompt Tuning in Computer Vision.}
Prompt Learning was initially introduced in the field of Natural Language Processing (NLP).
In GPT-2~\cite{radford2019language}, the pre-trained language model can complete specific downstream tasks without fine-tuning by adding some prefix descriptions, i.e., prompts, before the input sequence. 
Some works~\cite{schick2020exploiting,shin2020autoprompt} also make prompts learnable to better adapt to downstream tasks.
In general, the existing methods of prompt tuning in computer vision can be roughly categorized into two clusters: improving the discrimination abilities and enhancing the generalization abilities.
For the discrimination abilities, CoOp~\cite{zhou2022learning} first introduces prompt tuning of the CLIP into few-shot prediction while further works improve the performance on various tasks, including fine-grained object retrieval~\cite{wang2023fine}, multi-label recognition~\cite{guo2023texts} and long-tailed classification~\cite{dong2022lpt}.
For the generalization abilities, CoCoOp~\cite{zhou2022conditional} adapts to new target domains by making the prompt relevant to the input image while
ProGrad~\cite{zhu2022prompt} achieves it by gradient correction. More recently, KgCoOp~\cite{yao2023visual} propose to enhance generalization by constraining the learned prompt embeddings with general knowledge, and MaPLe~\cite{khattak2023maple} achieve it by learning coupled prompts in both visual and text modalities.
Aside from this, some works~\cite{zhang2021tip, zhang2023prompt} turn to retrieval in a knowledge base to achieve better performance, while CaFo~\cite{zhang2023prompt} \mbox{actually adopt $(K+K')$-shot for a $K$-shot problem.} 

However, those prompt tuning based methods need to utilize all category names to learn category centers before the text encoder. When the number of categories is large, this results in a significantly larger text batch, leading to substantial resource consumption. In contrast, we explore a text-guided prompt tuning paradigm that relocates the learning of category centers after the text encoder, which only requires a small text batch.

\section{Method}
\begin{figure*}[t]
	\centering
	\includegraphics[width=0.97\linewidth]{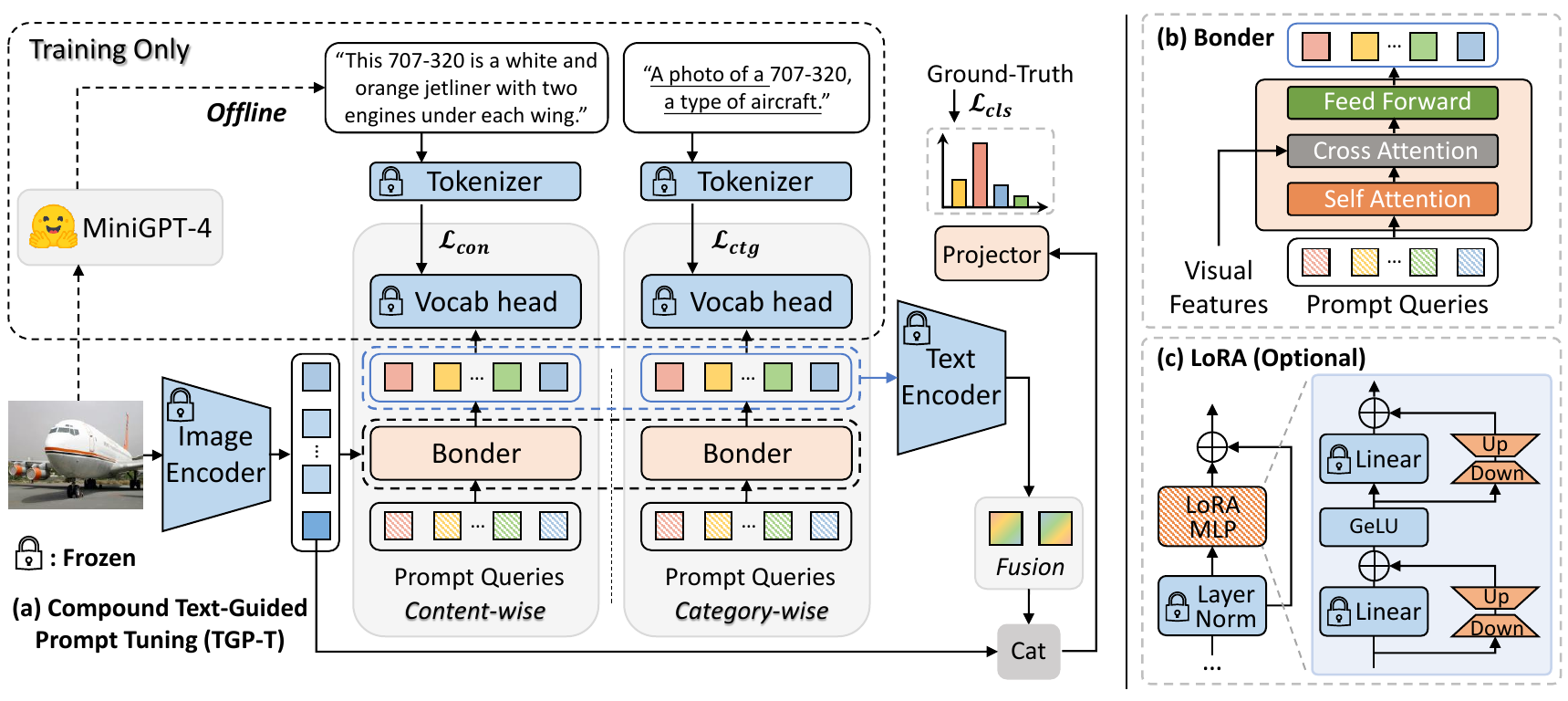}
         \caption{\textbf{(a)} Overview of the proposed TGP-T framework.
         1) Prompt Generation: the learnable prompts are conditioned on visual features through a Bonder structure.
         2) Text Supervision: we introduce content-wise and category-wise text supervision to guide the optimization of prompts during training.
         3) Feature Fusion: two text features are fused to yield the final text feature. The visual and text features are then concatenated and projected to perform recognition.
         Dashed lines denote that text descriptions are generated offline.
         \textbf{(b)} The detailed structure of Bonder.
         \textbf{(c)} TGP-T can further benefit from LoRA~\cite{hu2021lora}, where we identified tuning linear layers in MLP as an effective pattern. }
	\label{fig:framework}
\end{figure*}

Our method is based on a pre-trained vision-language model, as shown in Fig.~\ref{fig:framework}, and by adding a small number of learnable parameters, it allows for the cost-effective transfer of visual classification tasks.
In this section, we first give a brief review of CLIP~\cite{radford2021learning}. Then, we give a detailed introduction to the proposed TGP-T.

\subsection{Preliminary}
CLIP is an effective method to learn visual representation from natural language supervision.
Specifically, suppose the training set contains $M$ samples and is denoted as $\mathcal{S}=\{I_i, T_i\}_{i=0}^{M-1}$, where $I_i \in \mathbb{R}^{H\times W\times 3}$ is the image and $T_i$ is the textual description corresponding to the image $I_i$.
During training, the visual encoder $\mathcal{V}(\cdot)$ encodes the $I_i$ into visual feature: $\bm{v_i} = \mathcal{V}(I_i), \bm{v_i}\in \mathbb{R}^{d}$, where $d$ is the hidden dimension of CLIP. The textual encoder $\mathcal{T}(\cdot)$ encodes the $T_i$ into textual feature: $\bm{t_i} = \mathcal{T}(T_i), \bm{t_i}\in \mathbb{R}^{d}$.
Matched image and text feature pairs are regarded as positive pairs, i.e., $\{\bm{v_i}, \bm{t_i}\}$.
Correspondingly, unmatched pairs are regarded as negative pairs, i.e., $\{\bm{v_i}, \bm{t_j} | i \neq j\}$.
Given a batch of image-text pairs, CLIP maximizes the cosine similarity of positive pairs and minimizes it among negative pairs.
After such pre-training, CLIP can learn a good visual representation and be transferred to various downstream tasks.

Taking the classification task as an example, CLIP can accomplish zero-shot classification by reasonably constructing the text input.
CLIP utilizes a hand-crafted prompt to form the text input $T_i^{\prime}=\{\text{``A photo of a }[\text{CLASS}_i]\text{.''}\},i=0,1,...,N-1$,
where $[\text{CLASS}_i]$ is the specific category name, such as ``dog'', ``cat'', etc., and $N$ is the number of categories.
The image $I$ that needs to be predicted is fed into the image encoder to obtain the corresponding image feature: $\bm{v} = \mathcal{V}(I)$.
Then all $T_i^{\prime}$ is fed into the textual encoder in parallel to yield a set of textual features $\{\bm{t_i^{\prime}} | \bm{t_i^{\prime}} = \mathcal{T}(T_i^{\prime})\}_{i=0}^{N-1}$.
Based on the visual feature and textual feature, the probability is computed for CLIP's prediction:
\begin{equation}
\label{eq:pred_clip}
p(y=i|I)=\frac{e^{sim(\bm{t_{i}^{\prime}},\bm{v})/\tau}}{\sum_{i=0}^{N-1}{e^{sim(\bm{t_{i}^{\prime}},\bm{v})/\tau}}},
\end{equation}
where $sim(\cdot)$ denotes the similarity calculation and $\tau$ is a temperature parameter.
In our method, we directly \mbox{load a pre-trained CLIP model.}

\subsection{TGP-T: Compound Text-Guided Prompt Tuning}

\textbf{Prompt Generation.}
We condition the prompt generation on visual features $\bm{v}$ through a structure called Bonder, which is implemented with transformer layers, as shown in Fig.~\ref{fig:framework}.
Formally, We randomly initialize $K$ learnable prompt queries $\bm{Q}=\{\bm{q_1},\bm{q_2},...,\bm{q_K}\}\in \mathbb{R}^{K\times d}$. Then these queries are fed into Bonder, which can be formulated as follows:
\begin{equation}
\label{eq:bonder}
\begin{aligned}
&\bm{Q_{S}} = \bm{Q} + \text{Self-Attn}(\text{LN}(\bm{Q})),\\
&\bm{Q_{C}} = \bm{Q_{S}} + \text{Cross-Attn}(\text{LN}(\bm{Q_{S}}),\; \text{LN}(\bm{v})),\\
&\bm{P}= \bm{Q_{C}} + \text{FFN}(\text{LN}(\bm{Q_{C}})),\\
\end{aligned}
\end{equation}
where $\bm{P}=\{\bm{p_1},\bm{p_2},...,\bm{p_K}\}\in \mathbb{R}^{K\times d}$ is the generated prompts, $\text{Self-Attn}(\cdot)$ and $\text{Cross-Attn}(\cdot)$ denote the self-attention and cross-attention operation, respectively. $\text{LN}(\cdot)$ is the Layer Normalization and $\text{FFN}(\cdot)$ is a two-layer fully connected network.
Instead of concatenating category name tokens to the prompts, we directly feed $\bm{P}$ into the textual encoder.
The learnable $\bm{Q}$ will be updated during training through gradient backpropagation.
The interactions through Bonder insert plentiful visual cues and allow the prompts \mbox{to suit the current image better.}
\\
\\
\noindent\textbf{Text Supervision.}
The vanilla prompts generated above are short of textual semantics and lack specific category cues due to the absence of category names. 
Therefore, we introduce two ``teachers'' to guide the optimization of prompts from two distinct levels, i.e., content-wise and category-wise.
Accordingly, we construct two branches to learn two prompt inputs $\bm{P_{con}}$ and $\bm{P_{ctg}}$, respectively.  

As for content-wise text supervision, we take advantage of the prevailing VLMs to generate descriptions based on the specific content of the image.
Specifically, we employ MiniGPT-4~\cite{zhu2023minigpt}, which is powered by Vicuna-7B~\cite{vicuna2023} and is resource-efficient to generate descriptions: $D_{con}= \text{MiniGPT-4}(I)$.
We adopt the question ``\textit{Describe the \{class\} in this image in one sentence}''.
$D_{con}$ is generated offline before training, which ensures that no additional computational overhead is introduced \mbox{during the training process.}

As for the category-wise text supervision, we adopt the hand-engineered templates selected by Tip-Adapter~\cite{zhang2021tip},
which provides overall descriptions for the categories.
Take the FGVCAircraft dataset~\cite{maji2013fine} as an example, the text description is $D_{ctg}= \{\text{``A photo of a }[\text{CLASS}_i]\text{, a type of aircraft.''}\}_{i=0}^{N-1}$.
For more details of $\{D_{con}, D_{ctg}\}$ please refer to Appendix.

During the training process, we employ $D_{con}$ and $D_{ctg}$ to guide the optimization of $\bm{P_{con}}$ and $\bm{P_{ctg}}$, respectively.
Inspired by~\cite{devlin2018bert}, we calculate the loss in the vocabulary space $\mathbb{R}^{d_V}$, where $d_V$ denotes the vocabulary size of the pre-trained model. 
Specifically, we use the tokenizer of the pre-trained text encoder to map $\{D_{con}, D_{ctg}\}$ into the vocabulary space.
Then we project the generated prompts $\{\bm{P_{con}}, \bm{P_{ctg}}\}$ into the vocabulary space using the transposed weights of the pre-trained embedding layer, which is frozen and denoted as $\bm{W_E^T}\in \mathbb{R}^{d\times d_V}$.
The text supervision loss is then measured as follows:
\begin{equation}
\label{eq:loss_text}
\begin{aligned}
\{\mathcal{L}_{con},\mathcal{L}_{ctg}\}= \{&\text{CE}(\bm{P_{con}}\bm{W_E^T}, \text{Tokenizer}(D_{con})),\\
&\text{CE}(\bm{P_{ctg}}\bm{W_E^T}, \text{Tokenizer}(D_{ctg}))\},
\end{aligned}
\end{equation}
where $\mathcal{L}_{con}$ and $\mathcal{L}_{ctg}$ denote the loss for the content-wise and category-wise text supervision, respectively. $\text{CE}(\cdot)$ denotes the Cross-Entropy Loss.

\noindent\textbf{Cross-modal Feature Fusion.}
The $\{\bm{P_{con}}, \bm{P_{ctg}}\}$ generated through Bonder is then fed into text encoder to produce enriched features, which can be formulated as follows:
\begin{equation}
    \{ \bm{t_{con}}, \bm{t_{ctg}}\}=\{\mathcal{T}(\bm{P_{con}}), \mathcal{T}(\bm{P_{ctg}})\},
\end{equation}
where $\bm{t_{con}}\in \mathbb{R}^{d}, \bm{t_{ctg}} \in \mathbb{R}^{d}$ denote the textual features for content-wise and category-wise branches, respectively.

Subsequently, the text modality features and visual modality features are concatenated together and passed through a projector $\mathcal{F}(\cdot)$ to get the final predictions.
The classification loss is computed as follows:
\begin{equation}
    \mathcal{L}_{cls} = \text{CE}(\mathcal{F}([\bm{v}, (\bm{t_{con}} + \bm{t_{ctg}})/2]), y),
\end{equation}
where $[\cdot]$ donates the concatenated operation 
 along the feature dimension, and $\text{CE}(\cdot)$ donates the Cross-Entropy loss. $\mathcal{F}(\cdot)$ consists of a single linear layer followed by a softmax function, and $y$ is the corresponding label to the image $I$.
The fusion between text features follows a general practice of additive averaging.
\\
\\
\noindent\textbf{Model Training and Inference.}
During the training process, both the image encoder and text encoder are frozen, and only the Bonder and projector are learnable.
The overall loss can be formulated as follows:
\begin{equation}
    \mathcal{L} = \mathcal{L}_{cls} + \mathcal{L}_{con} + \mathcal{L}_{ctg}.
\end{equation}
In addition to directly freezing the weights of the pre-trained model, we also explored efficient fine-tuning methods such as LoRA~\cite{hu2021lora}, which is currently widely used for fine-tuning large language models.
Through our experiments, we have identified a reasonable way to apply it to visual tasks (i.e., tuning only linear layers in the MLP), ensuring that our method can also benefit from this fine-tuning approach.
For specific details, please refer to the Appendix.

During the inference stage, our method does not require any additional textual information, which is different from previous manual template-based methods and prompt tuning based frameworks.
We directly input the image into the network, and Bonder will adaptively generate suitable prompts to feed into the text encoder. Ultimately, the visual features and textual features are concatenated and projected to obtain the prediction results.

\section{Experiment}
\subsection{Experimental Settings}
\textbf{Datasets.}
Following CLIP~\cite{radford2021learning}, we adopt 11 publicly available image classification datasets that cover diverse scenes and scales, 
including ImageNet~\cite{deng2009imagenet}, Caltech~\cite{fei2004learning}, OxfordPets~\cite{parkhi2012cats}, Flowers~\cite{nilsback2008automated}, Food101~\cite{bossard2014food}, StanfordCars~\cite{krause20133d}, FGVCAircraft~\cite{maji2013fine}, EuroSAT~\cite{helber2019eurosat}, UCF101~\cite{soomro2012ucf101}, DTD~\cite{cimpoi2014describing}, and SUN397~\cite{xiao2010sun}.
We follow the few-shot evaluation protocol in CoOp~\cite{zhou2022learning}, i.e., we use 1, 2, 4, 8, and 16 shots for training, respectively, and report results on the full test sets.

\noindent\textbf{Implementation Details.}
We set ViT-B/16 as the image encoder.
The depth of the Bonder is set to 1.
The number of category-wise and content-wise prompt queries is 32 and 64, respectively.
We adopt the AdamW optimizer~\cite{loshchilov2017decoupled} with a learning rate of 5e-5 and a weight decay of 1e-4. 
The model is trained for 12,800 iterations with a batch size of 8. 
We tune the hyperparameters on a few-shot validation set with $min(n,4)$ shots ($n$ is the number of training shots) rather than searching on the test set.

\begin{table}[t]
    \centering
    \scalebox{1.0}{
        \small
        \begin{tabular}{p{47pt}<{\raggedright}|p{34pt}<{\centering}|p{14pt}<{\centering}p{14pt}<{\centering}p{14pt}<{\centering}p{15pt}<{\centering}p{16pt}<{\centering}}
        \toprule[1pt]
        \multirow{2}{*}{\hspace{-4pt}Method} & \multirow{2}{*}{Ref.} & \multicolumn{5}{c}{Number of Shots} \\
        \cline{3-7}
        & & \rule{0pt}{12pt} 1 & 2 & 4 & 8 & 16 \\ 
        \bottomrule[0.5pt]
        \rule{0pt}{9pt}\hspace{-5pt}Linear Probe & -- & 43.87 & 56.84 & 67.12 & 73.77 & 78.16 \\
        \hspace{-5pt}CoOp & IJCV'22& 68.39 & 71.50 & 74.45 & 77.03 & 79.86 \\
        \hspace{-5pt}CoCoOp & CVPR'22 & 69.10 & 70.38 & 72.32 & 76.20 & 78.43 \\
        \hspace{-5pt}Tip-Adapter & ECCV'22 & 69.81 & 71.56 & 74.18 & 75.17 & 77.39 \\
        \hspace{-5pt}Tip-Adapter-F & ECCV'22 & \underline{70.86} & 73.10 & 76.04 & 78.81 & 81.27 \\
        \hspace{-5pt}MaPLe & CVPR'23 & 69.93 & 72.54 & 76.37 & 79.02 & 81.79 \\
        \hspace{-5pt}Cross-Modal & CVPR'23 & 70.75 & 73.29 & 76.79 & 79.05 & 80.75 \\
        \rowcolor{Gray} \hspace{-5pt}TGP-T & -- & 70.51 & \underline{74.08} & \underline{77.13} & \underline{79.34} & \underline{82.65} \\
        \rowcolor{Gray} \rule{0pt}{7pt}\hspace{-5pt}TGP-T-F & -- & \textbf{72.15} & \textbf{75.22} & \textbf{78.20} & \textbf{80.69} & \textbf{84.06} \\
        \bottomrule[1pt]
        \end{tabular}
    }
    \caption{\textbf{Comparison (\%) to SOTA} using the CoOp protocol, which reports averaged top-1 accuracy across 11 test sets with ViT-B/16. 
    ``Linear Probe'' denotes the Linear-Probe CLIP.
    The best results are \textbf{bolded}, and the second best results are \underline{underlined}.}
    \label{tab:main_res}
\end{table}

\subsection{Performance} 
To evaluate the superiority of our novel framework, we compare with prior arts including CoOp~\cite{zhou2022learning}, CoCoOp~\cite{zhou2022conditional}, Cross-Modal Adaptation~\cite{lin2023multimodality}, MaPLe~\cite{khattak2023maple}, Tip-Adapter~\cite{zhang2021tip} with its fine-tuning version Tip-Adapter-F, and Linear-Probe CLIP~\cite{radford2021learning}.
We reproduce all previous methods using the same randomly sampled few-shot images for a fair comparison.
\\
\\
\noindent\textbf{Comparisons to Prior Arts.} As reported in Tab.~\ref{tab:main_res}, TGP-T surpasses previous methods from 2 shots to 16 shots, demonstrating its superior performance.
Remarkably, TGP-T with 2 shots outperforms the 4-shot CoCoOp and Linear-Probe CLIP. 
TGP-T with 8 shots performs better than 16-shot CoCoOp, Tip-Adapter, and Linear-Probe CLIP,
which underscores its effectiveness in learning from limited data.
Compared with recent work Cross-Modal, TGP-T achieves a 1.90\% increase in 16-shot settings.
Compared with CoCoOp, which conditions the prompt generation on visual features, TGP-T obtains an average gain from 78.43\% to 82.65\%.
Compared with Tip-Adapter, which constructs a knowledge base for categorization, TGP-T attains an average gain from 77.39\% to 82.65\%.
These results further demonstrate the superiority of our proposed framework.
Moreover, with LoRA~\cite{hu2021lora}, which fine-tunes a small number of parameters (less than 0.1\% of model parameters), TGP-T-F sets the new state-of-the-art performance across all shot settings, exceeding previous methods with a decent margin.
\begin{table}[t]
    \centering
    \scalebox{1.0}{
        \small
        \begin{tabular}{p{70pt}<{\raggedright}|p{38pt}<{\centering}p{0.05pt}<{\centering}p{30pt}<{\centering}p{30pt}<{\centering}}
        \toprule[1pt]
        \multirow{2}{*}{\hspace{-3pt}Method} & Source && \multicolumn{2}{c}{Target} \\
        \cline{2-2}\cline{4-5}
        & \rule{0pt}{12pt} ImageNet && -V2 & -Sketch \\
        \bottomrule[0.5pt]
        \rule{0pt}{9pt}\hspace{-3pt}Zero-Shot CLIP & 66.7 && 60.8 & 46.2 \\
        \hspace{-3pt}Linear Probe & 65.9 && 56.3 & 34.8 \\
        \hspace{-3pt}CoOp & 71.7 && 64.6 & 47.9 \\
        \hspace{-3pt}CoCoOp & 71.0 && 64.1 & 48.8 \\
        \hspace{-3pt}Cross-Modal & 72.8 && 64.8 & 47.9 \\
        \hspace{-3pt}MaPLe & 71.9 && 64.1 & \textbf{49.2} \\
        \rowcolor{Gray} \hspace{-3pt}TGP-T & \textbf{73.5} && \textbf{65.1} & 48.7 \\
        \bottomrule[1pt]
        \end{tabular}
    }
    \caption{\textbf{Comparison (\%) on Distribution Shift.} We train the model on ``Source'' dataset and test on ``Target'' datasets with ViT-B/16.}
    \label{tab:ood}
\end{table}
\\
\\
\noindent\textbf{Distribution Shift.}
We further assess the robustness of TGP-T under out-of-distribution (OOD) conditions.
Specifically, we train on the ImageNet~\cite{deng2009imagenet} and test on ImageNet-V2~\cite{recht2019imagenet} and ImageNet-Sketch~\cite{wang2019learning}.
As shown in Tab.~\ref{tab:ood}, when achieving the best result on the source dataset, TGP-T consistently attains promising performance on both OOD datasets.
It achieves the highest accuracy of 65.1\% on ImageNet-V2 and a competitive 48.7\% on the more challenging ImageNet-Sketch.
The results demonstrate the superior OOD performance of TGP-T. 
It effectively generalizes from ImageNet to the out-of-distribution datasets, showcasing its potential in handling distribution shifts.
\begin{table}[t]
    \centering
    \scalebox{1.0}{
        \small
        \begin{tabular}{p{54pt}<{\raggedright}|p{15pt}<{\centering}p{15pt}<{\centering}p{15pt}<{\centering}p{18pt}<{\centering}|p{14pt}<{\centering}p{20pt}<{\centering}}
        \toprule[1pt]
        \hspace{-4pt}Method & RN50 & B/32 & B/16 & L/14 & Mem. & Time \\ 
        \bottomrule[0.5pt]
        \rule{0pt}{10pt}\hspace{-5pt}Zero-Shot CLIP & 59.60 & 63.20 & 68.60 & 75.30 & - & - \\
        \hspace{-4pt}CoOp & 62.95 & 66.85 & 71.60 & \textcolor{red}{\textit{OOM}} & \textcolor{red}{18} & \textcolor{red}{12hr} \\
        \hspace{-4pt}CoCoOp & 63.30 & 66.20 & 71.30 & \textcolor{red}{\textit{OOM}} & \textcolor{red}{18} & \textcolor{red}{13hr} \\
        \hspace{-4pt}Tip-Adapter & 62.03 & 65.48 & 70.19 & 77.06 & 5 & - \\
        \hspace{-4pt}MaPLe & - & 66.80 & 71.90 & \textcolor{red}{\textit{OOM}} & \textcolor{red}{20} & 2hr \\
        \rowcolor{Gray} \rule{0pt}{10pt}\hspace{-4pt}TGP-T & \textbf{65.19} & \textbf{68.15} & \textbf{73.48} & \textbf{79.07} & \textcolor[RGB]{50,205,50}{1} & \textcolor[RGB]{50,205,50}{12min} \\
        \bottomrule[1pt]
        \end{tabular}
    }
    \caption{\textbf{Comparison (\%) of Different Visual Backbones.} For ViT models, we take suffixes such as ``B/32'' as their names for simplicity. We also report the GPU Memory consumption (GB) and the Training Time of the ViT-B/16.
    All results are conducted on the 16-shot ImageNet. }
    \label{tab:backbone}
\end{table}
\\
\\
\noindent\textbf{Different Visual Backbones.} We implement TGP-T with various visual encoders. 
Moreover, we report the GPU memory usage and training time of the ViT-B/16 backbone.
As shown in Tab.~\ref{tab:backbone}, TGP-T consistently achieves leading performance with different backbones, indicating our generalizability to network architectures.
As for training costs, both CoOp, CoCoOp, and MaPLe are memory and time-intensive, consuming over 18GB of GPU memory and taking a long time to train.
In contrast, TGP-T consumes only 1GB of GPU memory. 
Moreover, TGP-T enables the utilization of more powerful backbones such as ViT-L/14, while CoOp, CoCoOp, and MaPLe run into out-of-memory (OOM) problems on Nvidia RTX 3090.
In addition, TGP-T trains in a short period, demonstrating better efficiency.

\subsection{Ablation Studies}
\begin{figure}[t]
	\centering
	\includegraphics[width=0.99\linewidth]{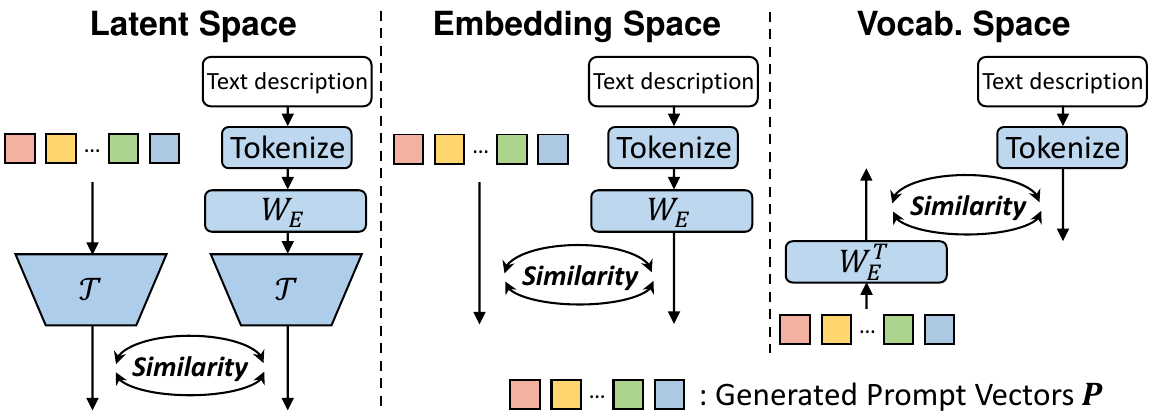}
	\caption{\mbox{Illustration of Different Spaces of Text Supervision.}}
	\label{fig:abl_supspace}
\end{figure}

We conduct analysis across all 11 datasets for the ablations on text supervision.
For other ablations, we evaluate on the most representative ImageNet.
\begin{figure}[t]
	\centering
	\includegraphics[width=0.83\linewidth]{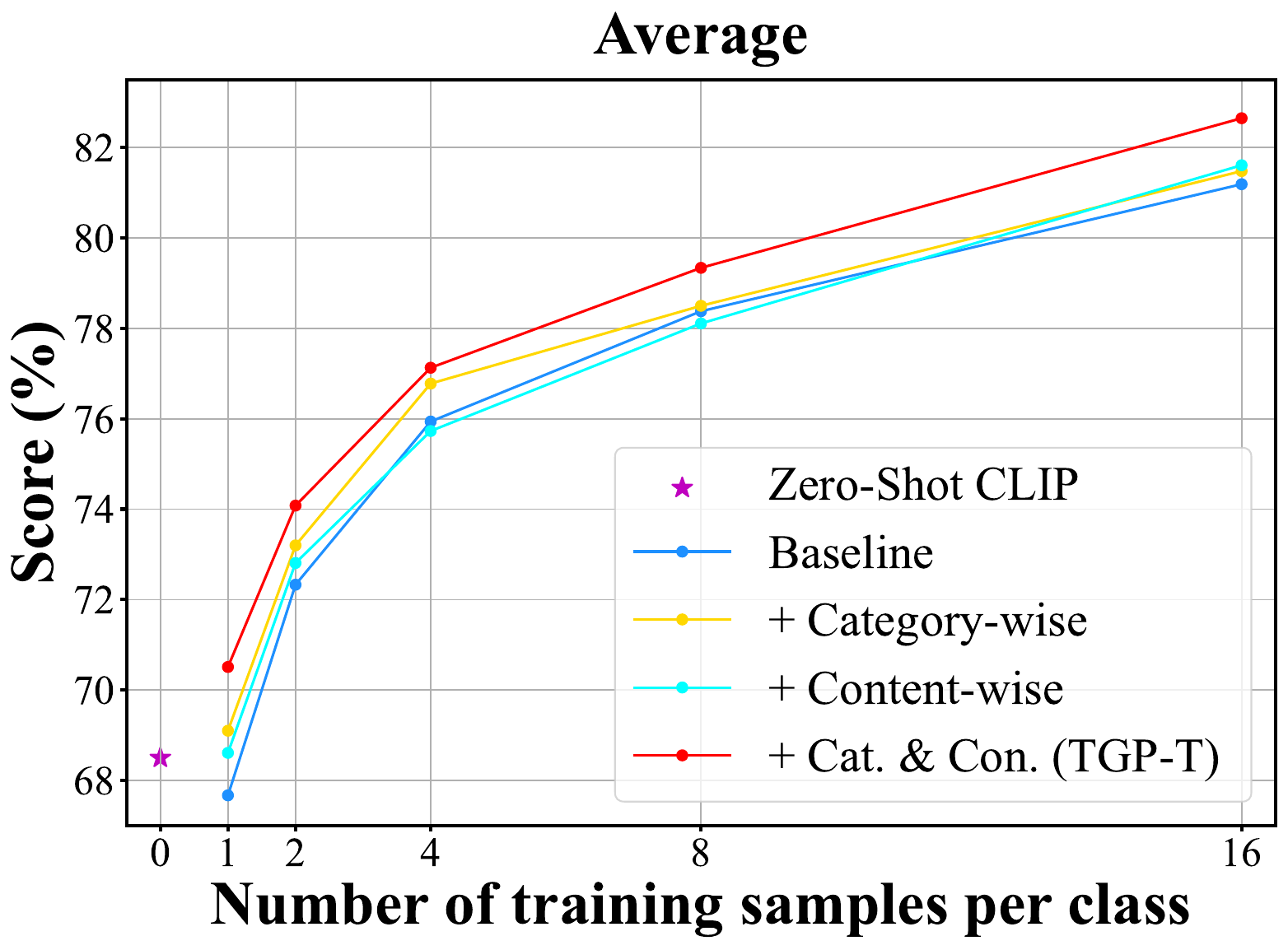}
	\caption{\textbf{Ablation Study of the Granularity of Text Supervision.} Results are averaged across 11 datasets. ``Baseline'' denotes the model trained without text supervision.}
	\label{fig:abl_textsup}
\end{figure}
\begin{table}[t]
    \centering
    \scalebox{1.0}{
        \small
        \begin{tabular}{p{15pt}<{\centering}p{17pt}<{\centering}|p{35pt}<{\centering}p{32pt}<{\centering}|p{32pt}<{\centering}p{32pt}<{\centering}}
        \toprule[1pt]
        \multirow{2}{*}{\rule{0pt}{11pt}Cat.} & \multirow{2}{*}{\rule{0pt}{11pt}Con.} & \multicolumn{2}{c|}{4-Shot} & \multicolumn{2}{c}{16-Shot} \\ 
        \cline{3-6}
        && \rule{0pt}{11pt}\hspace{-5pt} ImageNet & Aircraft & ImageNet & Aircraft \\
        \bottomrule[0.5pt]
        & & \rule{0pt}{11pt}70.18 & 32.10 & 72.82 & 42.63 \\
        \checkmark& & 70.06 & 34.53 & 73.02 & 47.49 \\
        &\checkmark & 70.41 & 33.60 & 73.09 & 47.40 \\
        \rowcolor{Gray} & & \textbf{70.58} & \textbf{36.60} & \textbf{73.48} & \textbf{52.39} \\
        \rowcolor{Gray} \multirow{-2}{*}{\checkmark} & \multirow{-2}{*}{\checkmark} & \textcolor[RGB]{50,205,50}{($\uparrow$0.40)} & \textcolor[RGB]{50,205,50}{($\uparrow$4.50)} & \textcolor[RGB]{50,205,50}{($\uparrow$0.66)} & \textcolor[RGB]{50,205,50}{($\uparrow$9.76)}\\
        \bottomrule[1pt]
        \end{tabular}
    }
    \caption{\textbf{Ablation Study (\%) of the Granularity of Text Supervision.} ``Cat.'' and ``Con.'' denote Category-wise and Content-wise text supervision, respectively.}
    \label{tab:abl_textsup}
\end{table}
\\
\\
\noindent\textbf{Granularity of Text Supervision.} 
We investigate the influence of different text supervision. 
As shown in Fig.~\ref{fig:abl_textsup}, the model trained without text supervision is denoted as ``Baseline''.
In general, concurrently employing both types of text supervision results in significant improvements.
Digging deeper, when the number of shots is small, using category-wise supervision alone yields comparable performance. 
As the number of shots increases, only combining both types of text supervision achieves a prominent lead. 
This is because when there is only 1 shot, category-wise description is sufficient to provide overall information. 
As the number of shots increases, intra-class diversity starts to emerge.
The standalone edition of category-wise and content-wise supervision can not significantly improve the performance, since each provides only a partial perspective.
\textit{Instead, combining both types of supervision helps the model understand the general category information while capturing intra-class variations, which leads to a noticeable improvement in categorization.}
In Tab.~\ref{tab:abl_textsup}, we further discuss results on the most representative ImageNet, and FGVCAircraft, where the pre-defined category names such as ``707-320'' are ambiguous for the model.
Combining both types of text supervision leads to 4.50\% and 9.76\% gains on FGVCAircraft with 4 and 16 shots, respectively, which verifies the effectiveness of compound text supervision.
\\
\\
\noindent\textbf{Different Spaces of Text Supervision.}
As illustrated in Fig.~\ref{fig:abl_supspace}, there are several spaces to employ text supervision.
The Embedding Space is after the embedding layer, and the Latent Space is after the text encoder.
The Vocabulary Space refers to the discrete representation space of words~\cite{devlin2018bert}.
As shown in Tab.~\ref{tab:abl_loss}, applying supervision in the vocabulary space leads to better performance across all shot settings. 
In fact, discrete vocabulary space is inherently more structured than continuous feature space. 
\textit{Guiding the optimization of prompts in discrete space reduces ambiguity and simplifies the learning process}, which further makes the task of distinguishing between different categories more straightforward.
Therefore, we suggest that incorporating discrete vocabulary space to guide prompt learning is more effective.
\begin{table}[t]
    \centering
    \scalebox{1.0}{
        \small
        \begin{tabular}{p{78pt}<{\centering}|p{16pt}<{\centering}p{16pt}<{\centering}p{16pt}<{\centering}p{16pt}<{\centering}p{16pt}<{\centering}}
        \toprule[1pt]
        \multirow{2}{*}{\rule{0pt}{8pt}\hspace{-4pt}Supervision Space} & \multicolumn{5}{c}{Number of Shots} \\
        \cline{2-6}
        \hspace{-4pt} & \rule{0pt}{10pt}1 & 2 & 4 & 8 & 16 \\ 
        \bottomrule[0.5pt]
        \hspace{-4pt}\rule{0pt}{10pt}Embedding Space & 68.59 & 72.50 & 76.07 & 77.66 & 81.46 \\
        \hspace{-4pt}Latent Space & 67.80 & 72.39 & 76.23 & 78.36 & 81.58 \\
        \rowcolor{Gray}\hspace{-4pt}Vocab. Space (TGP-T) & \textbf{70.51} & \textbf{74.08} & \textbf{77.13} & \textbf{79.34} & \textbf{82.65} \\
        \bottomrule[1pt]
        \end{tabular}
    }
    \caption{\textbf{Ablation Study (\%) of the Supervision Space.} Results are averaged across 11 datasets.}
    \label{tab:abl_loss}
\end{table}
\\
\\
\noindent\textbf{Different Structures of Bonder.}
There are several choices for the Bonder structure.
We evaluate the Self-Attention module, Cross-Attention module, and Meta-Net. 
The Meta-Net is adopted in CoCoOp~\cite{zhou2022conditional}, which denotes a two-layer bottleneck structure (Linear-ReLU-Linear).
As shown in Tab.~\ref{tab:bonder_structure}, the Cross-Attention consistently outperforms the others.
This is attributed to its ability to aggregate information from different sources, enabling an effective incorporation of visual cues.

\subsection{Further Analysis}
\textbf{Generalization to different Text Encoders.}
We assess the generalizability of TGP-T to different text encoders.
As shown in Tab.~\ref{tab:text_encoder_comparison}, the CLIP's text encoder achieves the best performance with 1 shot. 
When the training shots increase, the FLAN-T5$_{\text{Base}}$ delivers superior performance. Above all,

\begin{table}[H]
    \centering
    \scalebox{1.0}{
        \small
        \begin{tabular}{p{87pt}<{\centering}|p{14pt}<{\centering}p{15pt}<{\centering}p{15pt}<{\centering}p{15pt}<{\centering}p{16pt}<{\centering}}
        \toprule[1pt]
        \multirow{2}{*}{Bonder Design} & \multicolumn{5}{c}{Number of Shots} \\
        \cline{2-6}
        & \rule{0pt}{12pt} 1 & 2 & 4 & 8 & 16 \\ 
        \bottomrule[0.5pt]
        \rule{0pt}{9pt}\hspace{-5pt}Meta-Net & 67.34 & 69.78 & 69.74 & 70.59 & 72.67 \\
        \hspace{-5pt}Self-Attention & 68.82 & 69.60 & 69.72 & 71.28 & 72.48 \\
        \rowcolor{Gray}\hspace{-5pt}Cross-Attention (TGP-T) & \textbf{69.32} & \textbf{70.12} & \textbf{70.58} & \textbf{72.07} & \textbf{73.48} \\
        \bottomrule[1pt]
        \end{tabular}
    }
    \caption{\textbf{Different Structures of Bonder.} Results are reported on the most representative ImageNet dataset.}
    \label{tab:bonder_structure}
\end{table}
\begin{table}[H]
    \centering
    \scalebox{1.0}{
        \small
        \begin{tabular}{p{25pt}<{\centering}|p{50pt}<{\centering}|p{15pt}<{\centering}p{15pt}<{\centering}p{15pt}<{\centering}p{15pt}<{\centering}p{16pt}<{\centering}}
        \toprule[1pt]
        \multirow{2}{*}{\hspace{-4pt}Method} & \multirow{2}{*}{Text Encoder} & \multicolumn{5}{c}{Number of Shots} \\
        \cline{3-7}
        & & \rule{0pt}{12pt} 1 & 2 & 4 & 8 & 16 \\ 
        \bottomrule[0.5pt]
        \rule{0pt}{9pt}\hspace{-5pt}TGP-T & BERT$_{\text{Base}}$ & 68.99 & \textbf{70.13} & 70.80 & 72.15 & 73.55 \\
        \hspace{-5pt}TGP-T & FLAN-T5$_{\text{Base}}$ & 68.45 & 69.64 & \textbf{70.84} & \textbf{72.30} & \textbf{73.67} \\
        \hspace{-5pt}TGP-T & CLIP-Text & \textbf{69.32} & 70.12 & 70.58 & 72.07 & 73.48 \\
        \bottomrule[1pt]
        \end{tabular}
    }
    \caption{\textbf{Generalization to different Text Encoders.} Results are reported on the most representative ImageNet.}
    \label{tab:text_encoder_comparison}
\end{table}
\begin{table}[H]
    \centering
    \scalebox{1.0}{
        \small
        \begin{tabular}{p{60pt}<{\centering}|p{18pt}<{\centering}p{18pt}<{\centering}p{18pt}<{\centering}p{18pt}<{\centering}p{18pt}<{\centering}}
        \toprule[1pt]
        \multirow{2}{*}{Bonder Depth} & \multicolumn{5}{c}{Number of Shots} \\
        \cline{2-6}
        & \rule{0pt}{12pt} 1 & 2 & 4 & 8 & 16 \\ 
        \bottomrule[0.5pt]
        \rule{0pt}{9pt}\hspace{-5pt}$\times$1 & \textbf{69.32} & \textbf{70.12} & 70.58 & 72.07 & \textbf{73.48}  \\
        \hspace{-5pt}$\times$2 & 69.30 & 69.76 & 70.66 & \textbf{72.23} & 73.29  \\
        \hspace{-5pt}$\times$4 & 69.12 & 70.04 & \textbf{70.73} & 71.71 & 73.23  \\
        \hspace{-5pt}$\times$8 & 69.13 & 69.61 & 69.96 & 71.46 & 72.91  \\
        \bottomrule[1pt]
        \end{tabular}
    }
    \caption{\textbf{Effects of Bonder Depth.} Results are reported on the most representative ImageNet dataset.}
    \label{tab:bonder_depth}
\end{table}

\noindent{different} text encoders achieve competitive results, while slightly scaling up to FLAN-T5$_{\text{Base}}$, our method
achieves an improvement accordingly, demonstrating the adaptability of the TGP-T to different text encoders. 
\\
\\
\noindent\textbf{Effects of Bonder Depth.}
We evaluate the influence of the depth of Bonder.
As shown in Tab.~\ref{tab:bonder_depth}, a depth of 1 yields the most robust results across 1, 2, and 16 shots, while a deeper bonder brings improvements with 4 and 8 shots.
Interestingly, a depth of 8 underperforms across all shot settings, suggesting that an overly deep bonder leads to overfitting or optimization difficulties.
Therefore, we suggest that a depth of 1 is a better trade-off between performance and efficiency.

\section{Conclusions}
In this work, we propose TGP-T, an efficient prompt tuning framework for adapting VLMs with significantly lower resource demand.
We introduce compound text supervision to guide the optimization of prompts.
Through a Bonder structure, we align the generated prompts with visual features.
As a result, we only need two prompt inputs to text encoder to produce state-of-the-art performance on 11 datasets for few-shot classification.
Future works could explore more diverse forms and task-adaptive text supervision to further improve the effectiveness of text supervision in prompt tuning.

\section{Acknowledgments}
This work was supported by the National Key Research and Development Plan under Grant 2021YFE0205700, Beijing Natural Science Foundation JQ23016, the External cooperation key project of Chinese Academy Sciences  173211KYSB20200002, the Science and Technology Development Fund of Macau Project 0123/2022/A3, and 0070/2020/AMJ, Open Research Projects of Zhejiang Lab No. 2021KH0AB07, CCF-Zhipu AI Large Model OF 202219 and InnoHK program.

\bibliography{main}

\newpage
\appendix

\section{Appendix}

\subsection{A. Additional Performance Comparison}
\textbf{Per-dataset Performance.} In Tab.~\ref{tab:add_main_res}, we present the per-dataset results for all methods.
Our method outperforms prior methods on most datasets.
On several datasets such as DTD, EuroSAT, FGVCAircraft, and StanfordCars, we surpass previous methods by a decent margin across all shot settings.
Overall, we achieve a superior average performance.
\\
\\
\noindent\textbf{GPU Memory Consumption.} In Tab.~\ref{tab:add_gpu}, we report the per-dataset GPU memory consumption. The consumption of prior works is related to the number of categories. TGP-T consistently consumes less GPU memory, especially when the number of categories increases.
Note that CoCoOp, which also conditions prompts on visual features, exhibits an exponential increase in GPU memory consumption with the growth of batch size.
One can only access CoCoOp on ImageNet using batch size of 1 with Nvidia RTX 3090, which results in a significantly long training time.

\subsection{B. Additional Ablation Results}
\textbf{Granularity of Text Supervision.}
In Tab.~\ref{tab:add_abl_textsup}, we report the averaged results of the ablation on Granularity of Text Supervision.
When the number of shots increases, the standalone edition of category-wise and content-wise supervision may not achieve significant improvements, while employing both concurrently brings benefits.

\begin{table}[h]
    \centering
    \renewcommand\arraystretch{1.05}
    \scalebox{1.0}{
        \small
        \begin{tabular}{p{15pt}<{\centering}p{17pt}<{\centering}|p{24pt}<{\centering}p{24pt}<{\centering}p{24pt}<{\centering}p{24pt}<{\centering}p{24pt}<{\centering}}
        \toprule[1pt]
        \multirow{2}{*}{\rule{0pt}{11pt}Cat.} & \multirow{2}{*}{\rule{0pt}{11pt}Con.} & \multicolumn{5}{c}{Numbder of Shots} \\ 
        \cline{3-7}
        && \rule{0pt}{11pt} 1 & 2 & 4 & 8 & 16 \\
        \bottomrule[0.5pt]
        & & \rule{0pt}{10pt}67.67 & 72.33 & 75.94 & 78.38 & 81.19 \\
        \checkmark& & 69.10 & 73.20 & 76.78 & 78.50 & 81.48 \\
        &\checkmark & 68.61 & 72.81 & 75.73 & 78.11 & 81.61 \\
        \rowcolor{Gray} && \textbf{70.51} & \textbf{74.08} & \textbf{77.13} & \textbf{79.34} & \textbf{82.65} \\
        \rowcolor{Gray} \multirow{-2}{*}{\checkmark} & \multirow{-2}{*}{\checkmark} & \textcolor[RGB]{50,205,50}{($\uparrow$2.84)} & \textcolor[RGB]{50,205,50}{($\uparrow$1.75)} & \textcolor[RGB]{50,205,50}{($\uparrow$1.19)} & \textcolor[RGB]{50,205,50}{($\uparrow$0.96)} & \textcolor[RGB]{50,205,50}{($\uparrow$1.46)} \\
        \bottomrule[1pt]
        \end{tabular}
    }
    \caption{\textbf{Ablation Study (\%) of the Granularity of Text Supervision.}}
    \label{tab:add_abl_textsup}
\end{table}

\noindent\textbf{Discussions on LoRA.}
As for a pre-trained weight matrix $W_0 \in \mathbb{R}^{d \times k}$, we replace its update by a low-rank decomposition:
\begin{equation}
\label{eq:lora}
W_0+\Delta W=W_0+B A, 
\end{equation}
where $B \in \mathbb{R}^{d \times r}$, $A \in \mathbb{R}^{r \times k}$.
As a result, the number of learnable parameters in LoRA is related to $r\ (r\ll min(d,k))$. In our experiments, we set $r=4$, which results in only 0.01\% trainable parameters of backbone.

As LoRA is initially introduced in NLP, migrating to CV tasks needs careful design.
We first investigate which modules to use LoRA.
In Tab.~\ref{tab:add_abl_lora_modules}, we can observe that: (1) fine-tuning Attention and MLP layers concurrently brings performance drop while tuning MLP layers alone is more effective. (2) Tuning visual encoder is more important than the text encoder, while tuning both gains further improvement.

\begin{algorithm}[t]
\caption{Compound Text-Guided Prompt Tuning}
\label{alg:algorithm}
\textbf{Input}: Images $I$, Text Descriptions $\{D_{con},D_{ctg} \}$;\\
\textbf{Trainable Parameters}: Bonder $\bm{\theta_b}$, Projector $\bm{W}$, Prompt queries $\{\bm{Q_{con}}, \bm{Q_{ctg}} \}$, (Optional) LoRA layers $\bm{\theta_l}$;\\
\textbf{Output}: Updated parameters;
\begin{algorithmic}[1] 
\STATE Random initialize $\{\bm{Q_{con}}, \bm{Q_{ctg}} \}$. Initialize image encoder $\mathcal{V}$ and text encoder $\mathcal{T}$ from pre-trained weights;
\STATE Extract image features $[\bm{v},\bm{X}]$;
\FOR{$i=0,1,...,iters-1$}
\STATE Construct prompts $\{\bm{P_{con}}, \bm{P_{ctg}} \}$ through Bonder $\bm{\theta_b}$;
\STATE Impose text supervision on $\{\bm{P_{con}}, \bm{P_{ctg}} \}$ from $\{D_{con},D_{ctg} \}$. Get loss $\mathcal{L}_{con}$ and $\mathcal{L}_{ctg}$ in Eq. 3;
\STATE Feed $\{\bm{P_{con}}, \bm{P_{ctg}} \}$ into $\mathcal{T}$ to get textual features: $\bm{t_{con}} \leftarrow \mathcal{T}(\bm{P_{con}})$, $\bm{t_{ctg}} \leftarrow \mathcal{T}(\bm{P_{ctg}})$;
\STATE Get classification loss $\mathcal{L}_{cls}$ in Eq. 5;
\STATE Update $\{\bm{\theta_b}, \bm{W}, \bm{Q_{con}}, \bm{Q_{ctg}}\}$ under the final objective $\mathcal{L}$ in Eq. 6;
\IF {LoRA is True}
\STATE Update $\bm{\theta_l}$ under $\mathcal{L}$;
\ENDIF
\ENDFOR
\end{algorithmic}
\end{algorithm}

\noindent\textbf{Hyperparameters $r$.}
We evaluate the impact of $r$ in Eq.~\ref{eq:lora}.
As shown in Tab.~\ref{tab:add_abl_rank}, $r$ is not a crucial factor, and using 4 is a suitable choice.

\subsection{C. Additional Visualization}
\textbf{Examples of Text Descriptions.}
The category-wise templates of 11 datasets are presented in Tab.~\ref{tab:add_templates}.
Each dataset has one template that contains the scenario prior, except for ImageNet, which adopts seven templates.

\begin{figure}[H]
	\centering
	\includegraphics[width=1.0\linewidth]{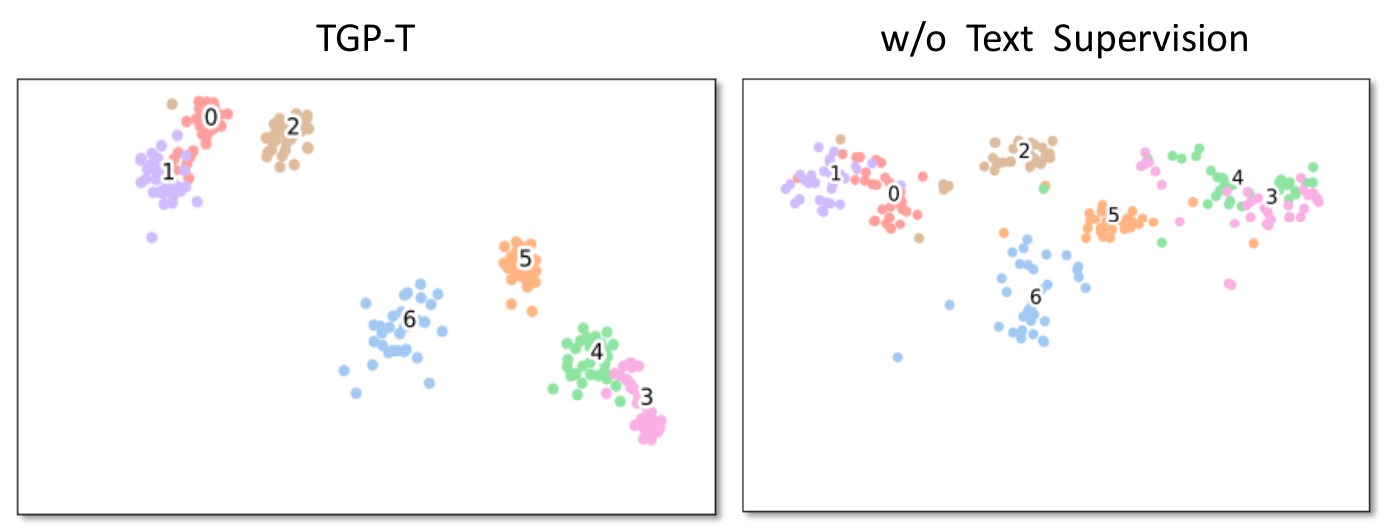}
        \caption{\textbf{t-SNE Visualization.} Different colors denote different categories on the test set of FGVCAircraft.}
	\label{fig:add_tsne}
 \vspace{9cm}
\end{figure}

\begin{figure*}[h]
	\centering
	\includegraphics[width=0.98\linewidth]{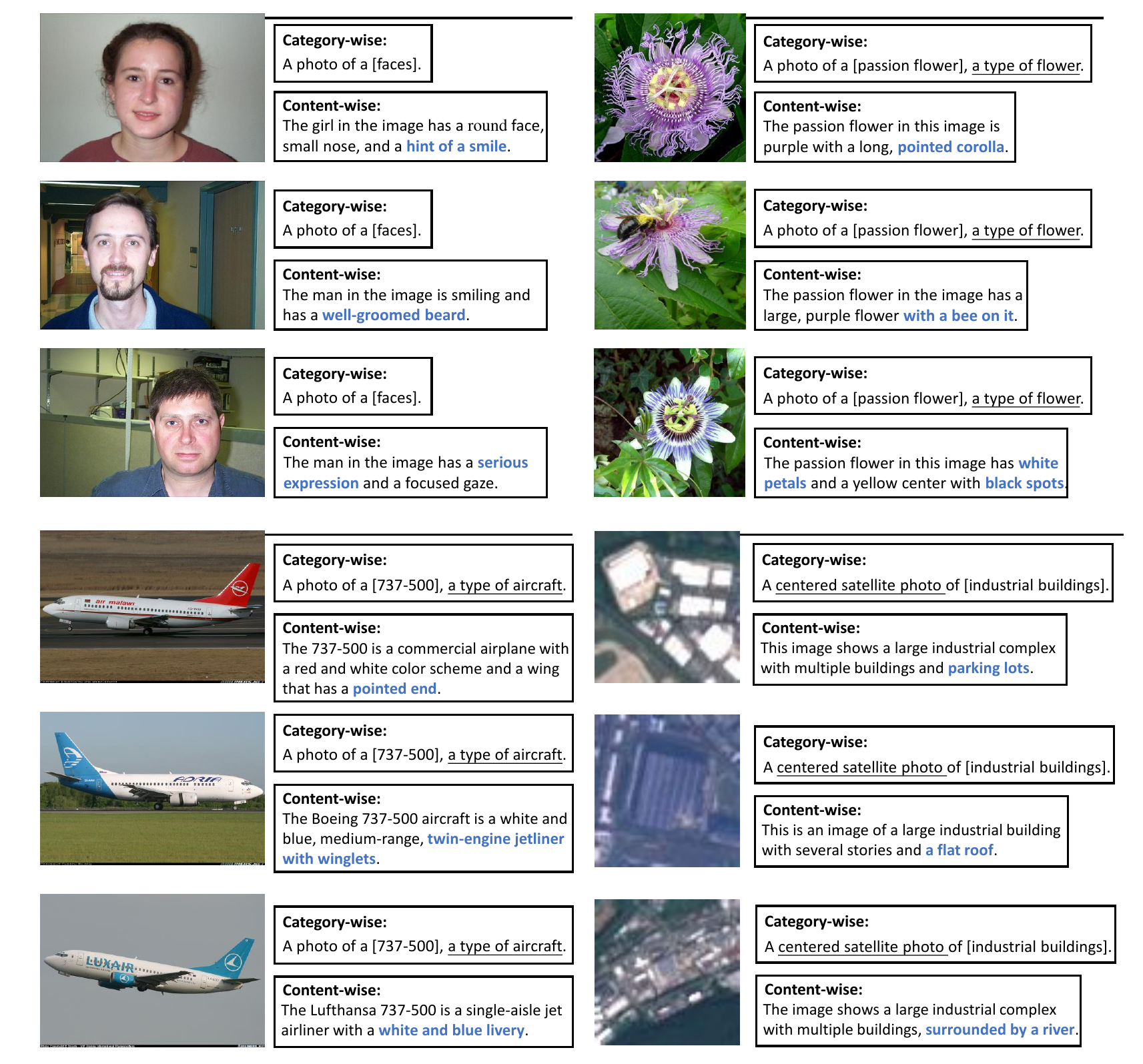}
	\caption{\textbf{Examples of Generated Text Descriptions.} Above examples are from Caltech, Flowers, FGVCAircraft, and EuroSAT dataset, respectively.}
	\label{fig:add_text_examples}
\end{figure*}

In Fig.~\ref{fig:add_text_examples}, we provide examples of generated text descriptions.
The texts labeled in blue indicate the distinct content of the sample, which may help capture intra-class variations.
\\
\\
\noindent\textbf{t-SNE.}
In Fig.~\ref{fig:add_tsne}, we present the t-SNE visualization of TGP-T compared to the model trained without Text Supervision.
Without text supervision, the model exhibits poor feature discriminability among categories, while TGP-T yields \textit{higher intra-class aggregation} and \textit{inter-class discretization}, demonstrating better discriminative abilities.
\\
\\
\noindent\textbf{Pseudo Code.}
The complete training procedure is summarized in Alg.~\ref{alg:algorithm}.
Note that we do not require any text input during the inference stage.

\begin{table}[h]
    \centering
    \renewcommand\arraystretch{1.05}
    \scalebox{1.0}{
        \small
        \begin{tabular}{p{76pt}<{\raggedright}|p{17pt}<{\centering}p{17pt}<{\centering}p{17pt}<{\centering}p{17pt}<{\centering}p{18pt}<{\centering}}
        \toprule[1pt]
        \multirow{2}{*}{\rule{0pt}{11pt}LoRA Modules} & \multicolumn{5}{c}{Numbder of Shots} \\ 
        \cline{2-6}
        & \rule{0pt}{11pt} 1 & 2 & 4 & 8 & 16 \\
        \bottomrule[0.5pt]
        \rule{0pt}{9pt}MLP$+$Attn. ($\mathcal{V}+\mathcal{T}$) & 68.50 & 68.56 & 69.47 & 69.82 & 71.07 \\
        MLP ($\mathcal{V}+\mathcal{T}$) & \textbf{69.84} & \textbf{70.50} & 71.30 & \textbf{72.85} & \textbf{74.39} \\
        MLP ($\mathcal{V}$) & 69.62 & 70.47 & \textbf{71.35} & 72.73 & 74.22 \\
        MLP ($\mathcal{T}$) & 68.44 & 70.33 & 70.83 & 72.15 & 73.57 \\
        
        \bottomrule[1pt]
        \end{tabular}
    }
    \caption{\textbf{Ablation Study (\%) of LoRA modules.} We report results on ImageNet.}
    \label{tab:add_abl_lora_modules}
\end{table}

\begin{table*}[t]
    \centering
    \renewcommand\arraystretch{1.0}
    \scalebox{1.0}{
        \small
        \begin{tabular}{p{60pt}<{\raggedright}|p{24pt}<{\centering}p{24pt}<{\centering}p{24pt}<{\centering}p{24pt}<{\centering}p{24pt}<{\centering}p{24pt}<{\centering}p{24pt}<{\centering}p{24pt}<{\centering}p{24pt}<{\centering}p{24pt}<{\centering}p{35pt}<{\centering}}
        \toprule[1pt]
        \multirow{2}{*}{\hspace{-4pt}Method} & \multicolumn{11}{c}{Dataset} \\
        \cline{2-12}
        & \rule{0pt}{10pt}EuroSAT & Pets & DTD & Caltech & Aircraft & Food & UCF & Flowers & Cars & SUN & ImageNet\\
        \hline
        \rule{0pt}{9pt}\hspace{-4pt} \# Categories & 10 & 37 & 47 & 100 & 100 & 101 & 101 & 102 & 196 & 397 & 1000 \\
        \hline
        \rule{0pt}{9pt}\hspace{-4pt}CoOp (bs=32)$^{\dag}$ & 756 & 1258 & 1420 & 2414 & 2414 & 2414 & 2414 & 2414 & 4108 & 7736 & 18560 \\
        \hspace{-4pt}CoOp (bs=8) & 742 & 1253 & 1409 & 2403 & 2403 & 2403 & 2403 & 2403 & 4093 & 7723 & 18545 \\
        \hspace{-4pt}CoCoOp (bs=8) & 2486 & 6404 & 7597 & 15213 & 15213 & 15213 & 15213 & 15213 & \textcolor{red}{\textit{OOM}} & \textcolor{red}{\textit{OOM}} & \textcolor{red}{\textit{OOM}} \\
        \hspace{-4pt}CoCoOp (bs=2) & 987 & 1983 & 2283 & 4220 & 4220 & 4220 & 4220 & 4220 & 7504 & 14560 & \textcolor{red}{\textit{OOM}} \\
        \hspace{-4pt}CoCoOp (bs=1)$^{\dag}$ & 740 & 1251 & 1404 & 2402 & 2402 & 2402 & 2402 & 2402 & 4094 & 7728 & 18564 \\
        \rowcolor{Gray}\hspace{-4pt}TGP-T (bs=8) & 1264 & 1265 & 1265 & 1266 & 1266 & 1266 & 1266 & 1266 & 1269 & 1270 & \textcolor[RGB]{50,205,50}{1296} \\
        \rowcolor{Gray}\hspace{-4pt}TGP-T (bs=1) & 738 & 738 & 739 & 740 & 740 & 740 & 740 & 740 & 743 & 752 & \textcolor[RGB]{50,205,50}{768} \\

        \bottomrule[1pt]
        \end{tabular}
        }
        \caption{\textbf{Facts about GPU Memory Consumption (GB).} We run the methods on a Nvidia RTX 3090. For a fair comparison, we test different settings of batch size (bs). $\dag$ denotes the batch size used in the official implementation. ``OOM'' denotes the out-of-memory problem. All results are based on ViT-B/16 backbone.}
    \label{tab:add_gpu}
\end{table*}
\begin{table}[H]
    \centering
    \renewcommand\arraystretch{1.05}
    \scalebox{1.0}{
        \small
        \begin{tabular}{p{40pt}<{\centering}|p{18pt}<{\centering}p{18pt}<{\centering}p{18pt}<{\centering}p{18pt}<{\centering}p{18pt}<{\centering}p{18pt}<{\centering}}
        \toprule[1pt]
        \rule{0pt}{8pt}$r$ & 1 & 2 & 4 & 8 & 16 & 32 \\ 
        \bottomrule[0.5pt]
        \rule{0pt}{10pt}TGP-T & 74.18 & 74.16 & \textbf{74.39} & 74.32 & 74.23 & 74.34 \\
        \bottomrule[1pt]
        \end{tabular}
    }
    \caption{\textbf{Analysis (\%) of Hyperparameter $r$.} We report results on 16-shot ImageNet.}
    \label{tab:add_abl_rank}
\end{table}
\begin{table}[H]
    \centering
    \renewcommand\arraystretch{1.08}
    \scalebox{1.0}{
        \small
        \begin{tabular}{p{70pt}<{\centering}|p{140pt}<{\centering}}
        \toprule[1pt]
        Dataset & Hand-crafted Template \\ 
        \bottomrule[0.5pt]
        \rule{0pt}{9pt} Caltech & a photo of a [class]. \\
        DTD & [class] texture. \\
        EuroSAT & a centered satellite photo of [class]. \\
        FGVCAircraft & a photo of a [class], a type of aircraft. \\
        Flowers & a photo of a [class], a type of flower. \\
        Food & a photo of [class], a type of food. \\
        OxfordPets & a photo of a [class], a type of pet. \\
        StanfordCars & a photo of a [class]. \\
        SUN397 & a photo of a [class]. \\
        UCF101 & a photo of a person doing [class]. \\
        \hline
        \multirow{7}{*}{\rule{0pt}{9pt}ImageNet} & itap of a [class]. \\
        & a bad photo of the [class]. \\
        & a origami [class]. \\
        & a photo of the large [class]. \\
        & a [class] in a video game. \\
        & art of the [class]. \\
        & a photo of the small [class]. \\
        \bottomrule[1pt]
        \end{tabular}
    }
    \caption{\textbf{Hand-engineered Templates.} We adopt these templates as the Category-wise text descriptions.}
    \label{tab:add_templates}
\end{table}

\begin{table*}[t]
    \scalebox{1.0}{
        \small
        \begin{tabular}{p{60pt}<{\centering}|p{16pt}<{\centering}|p{22pt}<{\centering}p{30pt}<{\centering}p{19pt}<{\centering}p{25pt}<{\centering}p{22pt}<{\centering}p{19pt}<{\centering}p{22pt}<{\centering}p{19pt}<{\centering}p{19pt}<{\centering}p{19pt}<{\centering}p{19pt}<{\centering}|p{25pt}<{\centering}}
        \toprule[1pt]
        \multirow{2}{*}{\hspace{-4pt}Method} & \multirow{2}{*}{Shots} & \multicolumn{11}{c}{Dataset} &  \\
        \cline{3-14}
        & & \rule{0pt}{9pt}Caltech & ImageNet & DTD & EuroSAT & Aircraft & Food & Flowers & Pets & Cars & UCF & SUN & Average\\
        \hline
        \rule{0pt}{9pt}Zero-Shot CLIP & 0 & 89.30 & 68.60 & 46.00 & 54.10 & 27.10 & 89.20 & 70.40 & 88.90 & 65.60 & 69.80 & 65.20 & 66.75 \\
        \hline
        \multirow{5}{*}{Linear Probing} & \rule{0pt}{9pt}1 & 79.19 & 33.36 & 38.24 & 48.26 & 20.37 & 47.86 & 73.29 & 49.96 & 35.32 & 52.50 & 42.21 & 47.32 \\
            & 2 & 87.22 & 45.96 & 48.40 & 57.00 & 25.98 & 64.17 & 85.55 & 58.41 & 52.28 & 64.24 & 55.03 & 58.57 \\
            & 4 & 87.22 & 45.96 & 48.40 & 57.00 & 25.98 & 64.17 & 85.55 & 58.41 & 52.28 & 64.24 & 55.03 & 58.57 \\
            & 8 & 93.39 & 61.36 & 67.08 & 76.79 & 40.38 & 80.28 & 95.90 & 83.65 & 74.72 & 78.64 & 70.42 & 74.78 \\
            & 16 & 95.01 & 65.91 & 72.28 & 85.70 & 46.75 & 83.58 & 97.89 & 88.09 & 82.73 & 81.76 & 73.80 & 79.41 \\
        \hline
        \multirow{5}{*}{CoOp} & \rule{0pt}{9pt}1 & 92.00 & 66.60 & 49.70 & 55.10 & 27.10 & 85.50 & 78.90 & 90.80 & 66.20 & 70.20 & 67.50 & 68.15 \\
            & 2 & 93.10 & 64.80 & 54.00 & 67.10 & 30.00 & 84.20 & 87.10 & 91.10 & 70.40 & 72.50 & 66.90 & 71.02 \\
            & 4 & 94.10 & 67.10 & 58.30 & 70.60 & 32.90 & 85.40 & 93.20 & 92.40 & 74.30 & 77.20 & 70.00 & 74.14 \\
            & 8 & 94.40 & 69.30 & 64.10 & 77.10 & 39.90 & 82.60 & 94.90 & 91.60 & 78.70 & 79.80 & 72.20 & 76.78 \\
            & 16 & 95.50 & 71.60 & 69.30 & 85.00 & 43.60 & 84.30 & 96.80 & 91.00 & 82.90 & 81.80 & 74.80 & 79.69 \\
        \hline
        \multirow{5}{*}{CoCoOp} & \rule{0pt}{9pt}1 & 93.30 & 69.50 & 53.30 & 58.60 & 28.20 & 84.90 & 73.80 & \textbf{91.80} & 67.50 & 70.60 & 68.50 & 69.10 \\
            & 2 & 94.00 & 69.80 & 52.20 & 61.80 & 29.70 & 86.00 & 77.20 & \textbf{91.90} & 68.10 & 74.00 & 69.40 & 70.38 \\
            & 4 & 94.80 & 70.50 & 57.40 & 64.00 & 31.90 & 86.30 & 82.80 & 92.60 & 69.30 & 75.30 & 70.60 & 72.32 \\
            & 8 & 94.70 & 70.90 & 63.60 & 76.40 & 36.70 & 86.30 & 93.40 & 92.70 & 73.00 & 77.80 & 72.70 & 76.20 \\
            & 16 & 95.30 & 71.30 & 68.50 & 81.00 & 38.70 & 87.10 & 95.60 & 93.00 & 76.70 & 81.50 & 74.00 & 78.43 \\
        \hline
        \multirow{5}{*}{Tip-Adapter} & \rule{0pt}{9pt}1 & 93.14 & 68.92 & 51.60 & 65.77 & 27.93 & 86.11 & 83.23 & 89.15 & 66.70 & 69.20 & 65.72 & 69.77 \\
            & 2 & 93.91 & 69.05 & 52.78 & 69.12 & 31.92 & 86.19 & 86.44 & 90.41 & 68.23 & 71.05 & 66.94 & 71.46 \\
            & 4 & 94.60 & 69.25 & 59.87 & 75.26 & 32.91 & \textbf{86.32} & 91.27 & 90.81 & 71.32 & 74.09 & 68.96 & 74.06 \\
            & 8 & 93.71 & 69.70 & 63.18 & 72.52 & 35.61 & 86.38 & 93.59 & 90.32 & 72.08 & 76.42 & 70.99 & 74.95 \\
            & 16 & 95.05 & 70.19 & 66.13 & 78.40 & 39.84 & 86.48 & 94.52 & 91.88 & 75.24 & 78.30 & 72.00 & 77.09 \\
        \hline
        \multirow{5}{*}{Tip-Adapter-F} & \rule{0pt}{9pt}1 & 93.75 & 69.30 & 53.66 & 64.15 & 28.74 & \textbf{86.12} & 86.93 & 90.87 & 67.58 & 73.70 & 64.59 & 70.85 \\
            & 2 & 94.12 & 69.91 & 56.03 & 70.68 & 32.49 & \textbf{86.22} & 89.48 & 91.25 & 70.99 & 75.97 & 66.75 & 73.08 \\
            & 4 & 94.73 & 70.70 & 61.47 & 78.07 & 36.33 & 86.09 & 92.65 & 91.93 & 74.54 & 79.67 & 70.41 & 76.05 \\
            & 8 & 94.60 & 72.09 & 67.79 & 83.83 & 41.79 & 86.67 & 94.32 & 91.80 & 77.88 & 82.21 & 73.92 & 78.81 \\
            & 16 & 95.58 & 73.29 & 72.10 & 88.46 & 44.25 & 87.24 & 96.06 & 92.75 & 83.65 & 84.14 & 76.31 & 81.26 \\
        \hline
        \multirow{5}{*}{MaPLe} & \rule{0pt}{9pt}1 & 91.60 & 69.50 & 51.70 & 71.60 & 27.40 & 81.20 & 82.60 & 89.00 & 66.70 & 71.30 & 66.60 & 69.93 \\
            & 2 & 94.10 & 70.30 & 53.30 & 74.30 & 29.10 & 81.70 & 89.80 & 89.90 & 70.10 & 74.10 & 71.20 & 72.54 \\
            & 4 & 93.60 & 70.50 & 61.40 & 83.90 & 34.80 & 81.80 & 94.20 & 92.40 & 75.40 & 79.10 & 73.00 & 76.37 \\
            & 8 & 94.80 & 71.40 & 66.40 & \textbf{89.90} & 42.70 & 83.10 & 95.50 & 93.00 & 78.40 & 81.00 & 73.00 & 79.02 \\
            & 16 & 96.20 & 71.90 & 71.90 & 92.60 & 48.00 & 85.30 & 97.50 & 92.30 & 84.00 & 84.50 & 75.50 & 81.79 \\
        \hline
        \multirow{5}{*}{Cross-Modal} & \rule{0pt}{9pt}1 & 93.14 & 68.14 & 51.42 & 64.84 & 29.55 & 85.61 & 87.62 & 89.21 & 65.86 & \textbf{73.91} & 68.92 & 70.75 \\
            & 2 & 93.30 & 68.08 & 56.50 & 68.22 & \textbf{33.84} & 84.65 & 91.03 & 91.23 & 71.81 & 76.24 & 71.26 & 73.29 \\
            & 4 & 95.37 & 69.51 & 62.35 & 79.63 & 35.55 & 85.61 & 94.76 & 92.50 & 75.26 & 80.36 & 73.84 & 76.79 \\
            & 8 & 94.93 & 71.46 & 67.43 & 83.54 & 42.18 & 86.38 & 96.18 & 92.37 & 77.76 & 83.00 & 74.32 & 79.05 \\
            & 16 & 96.26 & 72.78 & 72.52 & 82.86 & 44.40 & 86.47 & 97.28 & 92.70 & 83.05 & 84.00 & 75.98 & 80.75 \\
        \hline
         \rowcolor{Gray}& \rule{0pt}{9pt}1 & 93.71 & 69.32 & 52.42 & 59.47 & 29.94 & 84.48 & \textbf{87.74} & 88.31 & 69.27 & 72.09 & 68.85 & 70.51 \\
            \rowcolor{Gray}& 2 & \textbf{94.73} & 70.12 & 58.27 & 71.93 & 32.49 & 85.83 & 90.34 & 91.66 & 72.29 & \textbf{76.29} & 70.97 & 74.08 \\
            \rowcolor{Gray}& 4 & 95.29 & 70.58 & 62.12 & 81.75 & 36.60 & 86.05 & \textbf{95.41} & 92.23 & 75.72 & 79.12 & 73.58 & 77.13 \\
            \rowcolor{Gray}& 8 & \textbf{95.50} & 72.07 & 69.21 & 83.67 & 41.64 & 86.87 & 96.59 & 92.23 & 78.42 & 81.73 & 74.83 & 79.34 \\
            \rowcolor{Gray} \multirow{-5}{*}{TGP-T}& 16 & 96.15 & 73.48 & 73.23 & 88.80 & 52.39 & 87.28 & \textbf{98.34} & 92.78 & 85.92 & 84.46 & 76.27 & 82.65 \\
        \hline
         \rowcolor{Gray}& \rule{0pt}{9pt}1 & \textbf{94.20} & \textbf{69.84} & \textbf{54.73} & \textbf{72.96} & \textbf{30.36} & 84.49 & 85.67 & 88.28 & \textbf{70.44} & 73.35 & \textbf{69.30} & \textbf{72.15} \\
            \rowcolor{Gray}& 2 & 93.91 & \textbf{70.50} & \textbf{60.28} & \textbf{80.32} & 33.00 & 85.92 & \textbf{91.27} & 91.33 & \textbf{73.39} & 75.41 & \textbf{72.13} & \textbf{75.22} \\
            \rowcolor{Gray}& 4 & \textbf{95.70} & \textbf{71.30} & \textbf{66.25} & \textbf{84.02} & \textbf{38.11} & 85.95 & 94.48 & \textbf{92.67} & \textbf{76.21} & \textbf{81.31} & \textbf{74.16} & \textbf{78.20} \\
            \rowcolor{Gray}& 8 & 95.21 & \textbf{72.85} & \textbf{69.38} & 87.23 & \textbf{43.86} & \textbf{87.01} & \textbf{97.40} & \textbf{93.11} & \textbf{81.71} & \textbf{83.98} & \textbf{75.86} & \textbf{80.69} \\
           \rowcolor{Gray} \multirow{-5}{*}{TGP-T-F} & 16 & \textbf{96.35} & \textbf{74.39} & \textbf{75.65} & \textbf{93.62} & \textbf{54.55} & \textbf{87.42} & 97.73 & \textbf{94.03} & \textbf{86.85} & \textbf{86.62} & \textbf{77.42} & \textbf{84.06} \\
        \bottomrule[1pt]
        \end{tabular}
        }
    \caption{\textbf{Per-dataset results on ViT-B/16 backbone.} For a fair comparison, we re-run all previous methods with our randomly sampled few-shot images. 
    The best results are \textbf{bolded} for each shot and each dataset.}
    \label{tab:add_main_res}
\end{table*}

\newpage

\end{document}